\documentclass{article}

\usepackage[final]{corl_2020} 
\usepackage{graphicx}
\usepackage{amsmath} 
\usepackage{amssymb}  
\usepackage{subcaption}
\usepackage{float}
\usepackage{siunitx}
\usepackage{xcolor}
\usepackage{pifont}
\usepackage{xcolor,colortbl}
\usepackage{booktabs}

\definecolor{somegray}{rgb}{0.5, 0.5, 0.5}
\newcommand{\darkgrayed}[1]{\textcolor{somegray}{#1}}
\makeatletter
\newcommand*\titleheader[1]{\gdef\@titleheader{#1}}
\AtBeginDocument{%
  \let\st@red@title\@title
  \def\@title{%
    \vskip-3em
    \bgroup\normalfont\large\centering\@titleheader\par\egroup
    \vskip1.5em\st@red@title}
}
\makeatother

\titleheader{ \darkgrayed{This paper has been accepted for publication at the
Conference on Robot Learning, Cambridge MA, USA. 
2020. } }

\title{Flightmare: A Flexible Quadrotor Simulator}

\definecolor{Gray}{gray}{0.85}
\definecolor{LightCyan}{rgb}{0.88,1,1}
\newcolumntype{a}{>{\columncolor{Gray}}c}
\newcolumntype{b}{>{\columncolor{white}}c}

\newcommand{\cmark}{\ding{51}}%
\newcommand{\xmark}{\ding{55}}%

\author{
Yunlong Song, Selim Naji, Elia Kaufmann, Antonio Loquercio, Davide Scaramuzza\\ \\
Robotics and Perception Group \\
Depts. Informatics and Neuroinformatics\\ 
University of Zurich and ETH Zurich\\
}

\begin{document}
\maketitle


\begin{abstract}

State-of-the-art quadrotor simulators have a rigid and highly-specialized structure: either are they really fast, physically accurate, or photo-realistic.
%
In this work, we propose a novel quadrotor simulator: Flightmare. 
Flightmare is composed of two main components: a configurable rendering engine built on Unity and a flexible physics engine for dynamics simulation.
Those two components are totally decoupled and can run independently of each other. 
This makes our simulator extremely fast: rendering achieves speeds of up to 230~Hz, while physics simulation of up to 200,000~Hz on a laptop.
In addition, Flightmare comes with several desirable features: (i) a large multi-modal sensor suite, including an interface to extract the 3D point-cloud of the scene; (ii) an API for reinforcement learning which can simulate hundreds of quadrotors in parallel; and (iii) integration with a virtual-reality headset for interaction with the simulated environment.
We demonstrate the flexibility of Flightmare by using it for two different robotic tasks: quadrotor control using deep reinforcement learning and collision-free path planning in a complex 3D environment.
\\
\\
\textbf{Website:} \url{https://uzh-rpg.github.io/flightmare} \\
\textbf{Code:} \url{https://github.com/uzh-rpg/flightmare}
\end{abstract}

\keywords{Quadrotor Simulator, Photo-realistic Rendering. }

\section{Introduction}
\label{sec: introduction}
Simulators are invaluable tools for the robotics researcher.
They allow developing and testing algorithms in a safe and inexpensive manner, without having to worry about the time-consuming and expensive process of dealing with real-world hardware.
The ideal simulator is: (i) \emph{fast}, to collect a large amount of data with limited time and compute; (ii) \emph{physically-accurate}, to represent the dynamics of the real world with high-fidelity; and (iii) \emph{photo-realistic}, to minimize the discrepancy between simulated and real-world sensors' observations. 
Those objectives are generally conflicting in nature: for example, the more a simulation is realistic, the slower it is.
Therefore, achieving all those objectives in a single monolithic simulator is challenging.

The landscape of currently available simulators is fragmented: some are extremely fast, \emph{e.g.} Mujoco~\citep{todorov2012mujoco}, while others have either really accurate dynamics~\citep{raisim, furrer2016rotors} or highly photo-realistic rendering~\citep{guerra2019flightgoggles}. 
One of the main limitations of those simulators is their rigid nature.
%
Specifically, they entrust the simulator developers, and not the end-users, to trade-off accuracy for speed.
%
However, this paradigm leaves some questions open:
What if we want to \emph{dynamically} change the underlying physics model? What if we want to \emph{actively} trade-off photo-realism for speed?
In this work, we answer these questions in the context of quadrotor simulation.
To do so, we propose \emph{Flightmare}, a new flexible simulator which puts the speed vs. accuracy trade-off in the hands of the end-users.

Flightmare is composed of two main blocks: a rendering engine, based on Unity~\citep{juliani2018unity}, and a physics model.
These blocks are completely decoupled and can run independently from each other.
Besides, each block is flexible by design.
Indeed, the rendering block can be used within a wide range of 3D realistic environments and generate visual information from low to high photo-realism.
With minimal additional computational costs, it is also possible to simulate sensor noise, \emph{e.g.} motion-blur, environment dynamics, \emph{e.g.} wind, and lens distortions~\citep{juliani2018unity}.
Similarly, the physics block offers full control to the user in terms of the desired robot dynamics and associated sensing.
Depending on the application, the users can easily switch between a basic (noise-free) quadrotor model and a more advanced rigid-body dynamics, including friction and rotor drag, or directly use the real platform dynamics like~\citep{guerra2019flightgoggles}.
Inertial sensing and motor encoders, which directly depend on the physics model, can also be noise-free or include different degrees of noise~\citep{furrer2016rotors, kohlbrecher2013hector}.

\begin{table}[tp]
	\centering
	\resizebox{\textwidth}{!}{%
        \begin{tabular}{@{}l|c|c|c|c|c|c|c@{}}
        	\toprule
        	\\[-0.4em]
        	Simulator & Rendering & Dynamics & Sensor Suite & Point Cloud & VR Headset & RL API & Vehicles\\ \\[-0.4em]
        	\hline
        	\hline
        	\\[-0.4em]
        	Hector~\citep{kohlbrecher2013hector} & OpenGL & Gazebo-based & IMU, RGB &    \textcolor{red}{\xmark} &  \textcolor{red}{\xmark}  &  \textcolor{red}{\xmark} & Single\\ \\[-0.4em]
        	RotorS~\citep{furrer2016rotors} & OpenGL & Gazebo-based & IMU, RGB, Depth &  \textcolor{red}{\xmark}   &  \textcolor{red}{\xmark} &  \textcolor{red}{\xmark} & Single\\ \\[-0.4em]
        	FlightGoggles~\citep{guerra2019flightgoggles} & Unity & Flexible & IMU, RGB &  \textcolor{red}{\xmark} &   \textcolor{green}{\cmark}  &  \textcolor{red}{\xmark} & Single\\ \\[-0.4em]
        	AirSim~\citep{shah2018airsim} & Unreal Engine & PhysX  & IMU, RGB, Depth, Seg &  \textcolor{red}{\xmark} &  \textcolor{green}{\cmark}  &  \textcolor{red}{\xmark} & Multiple \\ \\[-0.4em]
        	\hline
        	\\[-0.4em]
        	\textbf{Flightmare} & \textbf{Unity} & \textbf{Flexible} & \textbf{IMU, RGB, Depth, Seg} & \textcolor{green}{\cmark} &
        	\textcolor{green}{\cmark} &  \textcolor{green}{\cmark} & \textbf{Multiple} \\ \\[-0.4em]
        	\bottomrule
        \end{tabular}
        }
	\caption{A comparison of Flightmare to other open-source quadrotor simulators.}
\label{tab:simulators}
\end{table}

Apart from photo-realistic rendering and fast quadrotor dynamics simulation,
Flightmare comes with several desirable features with respect to currently available quadrotor simulators.
%
In contrast to FlightGoggles~\citep{guerra2019flightgoggles}, we provide interfaces to the popular robotics simulator Gazebo along with different high-performance physics engines. 
In contrast to AirSim~\citep{shah2018airsim}, we decouple the rendering module from the physics engine, which offers fast and accurate physics simulation when rendering is not required. 
%
Additionally,  Flightmare (i) can simulate up to several hundreds of agents in parallel, which is not only useful for multi-drone applications but also enables extremely fast data collection and training, which is crucial, e.g., for developing deep reinforcement learning (RL) applications, (ii) provides a standard wrapper (OpenAI Gym)~\citep{brockman2016gym} to several RL tasks, together with popular OpenAI baselines~\citep{stable-baselines} for state-of-the-art RL algorithms, and (iii) offers a rich and configurable sensor suite, together with an API to extract the full 3D information of the environment, in the form of a point-cloud.
Table~\ref{tab:simulators} summarizes the main differences between ours and other quadrotor simulators.

We quantitatively evaluate the speed of the rendering and dynamics blocks of Flightmare under a wide range of settings.
This study shows that we can achieve speeds of up to 230~Hz for the rendering block and up to 200,000~Hz for the dynamics block with a commodity multi-core laptop CPU.
In addition, we demonstrate the generality of our simulator by using it on two challenging robotics tasks: learning a sensorimotor control policy for a quadrotor, possibly subject to sensor failures; and path-planning in a complex 3D environment.
Those two tasks put completely different requirements on the simulation stack, which Flightmare can provide given its flexible interface.
However, our simulator is not limited to these problems and can be applied to a wide range of tasks.



\section{Related Work}
\label{sec: related work}
We review several existing open-source simulators that have been widely used by robotics and machine learning researchers. 
We highlight important features and limitations of each simulator.
We drew inspiration from the successes of previous work to design a flexible quadrotor simulator 
that combines their desirable features while addressing their limitations.

\textbf{RotorS and Hector:}
Both RotorS~\citep{furrer2016rotors} and Hector~\citep{kohlbrecher2013hector} are popular Micro Aerial Vehicle~(MAV) simulators built on Gazebo~\citep{koenig2004design}, which is a general robotic simulation platform and generally used with the popular Robot Operating System~(ROS).
Hector is a collection of open-source modules and is primarily used for autonomous mapping and navigation with rescue robots. 
RotorS provides several multi-rotor helicopter models such as the AscTec Hummingbird, Pelican, and Firefly.
These gazebo-based simulators have the capability of accessing multiple high-performance physics engines and simulating various sensors, ranging from laser range finders to RGB cameras. 
In particular, RotorS has been extensively used in robotics to develop algorithms on MAVs, such as drone racing~\citep{loquercio2019deep}, exploration~\citep{cieslewski2017rapid}, path-planning~\citep{nguyen2018real}, or mapping~\citep{hinzmann2018mapping}. 
Nevertheless, Gazebo has limited rendering capabilities and is not designed for efficient parallel dynamics simulation, which makes it difficult to develop learning-based systems. 

\textbf{AirSim and CARLA:} 
Both AirSim~\footnote{AirSim also has an experimental Unity release.}~\citep{shah2018airsim} and CARLA~\citep{Dosovitskiy17} are open-source photo-realistic simulators for autonomous vehicles built on Unreal Engine.
CARLA is mainly made for autonomous driving research and only provides the dynamics of ground vehicles.
Conversely, AirSim offers an interface to configure multiple vehicle models for quadrotors and supports hardware-in-the-loop~(HITL) as well as software-in-the-loop~(SITL) with flight controllers such as PX4. 
The vehicle is defined as a rigid body whose dynamics model is simulated using NVIDIA’s physics engine PhysX, a popular physics engine used by the large majority of today's video games.  
However, this physics engine is not specialized for quadrotors (or robots),  and it is tightly coupled with the rendering engine to allow simulating environment dynamics.
Because of this rigid connection between rendering and physics simulation, AirSim can achieve only limited simulation speeds.
This limitation makes it difficult to apply the simulator to challenging model-free reinforcement learning tasks, \emph{e.g.} training an end-to-end control policy for quadrotor stabilization under harsh initialized poses~\citep{rl_quadrotor} or flying through a fast moving gate~\citep{Yunlong2020}.


\textbf{FlightGoggles:} 
FlightGoggles~\citep{guerra2019flightgoggles} is a photo-realistic sensor simulator for perception-driven robotic vehicles. 
FlightGoggles consists of two separate components: a photo-realistic rendering engine built on Unity 
and a quadrotor dynamics simulation implemented in C++.
In addition, it also provides an interface with real-world vehicles and actors in a motion capture system.
FlightGoggles is very useful for rendering camera images given trajectories and inertial measurements from flying vehicles in real-world, 
in which the collected dataset~\citep{antoniniIJRRblackbird} is used for testing vision-based algorithms.
Flightmare shares the same motivation with FlightGoggles of decoupling the dynamics modeling from the photo-realistic rendering engine. 
However, our simulator offers a larger suite of sensors observations, an API to extract the environment point-cloud, and a more structured physics interface, which allows simulating multiple agents in parallel.
These characteristics open up several new opportunities for both robotics and machine learning research (see Section~\ref{sec: exp}).

Apart from the aforementioned simulators, there are many more existing simulators that have been widely adopted 
by other research communities. 
For example, MuJoCo~\citep{todorov2012mujoco} has been widely used by the reinforcement learning community for benchmark comparisons. 
Similarly, RaiSim~\citep{raisim} is a physics engine for robotics and AI research written in C++, 
that supports massive parallel dynamics simulation.
However, both simulators do not support complex 3D environments and photo-realistic image rendering. 
Sim4CV~\citep{muller2018sim4cv} is a photo-realistic simulator but made solely for computer vision applications.

\section{Methodology}
\label{sec: methodology}

\subsection{System Overview}
\begin{figure}[!htp]
    \centering
    \includegraphics[width=1.0\textwidth]{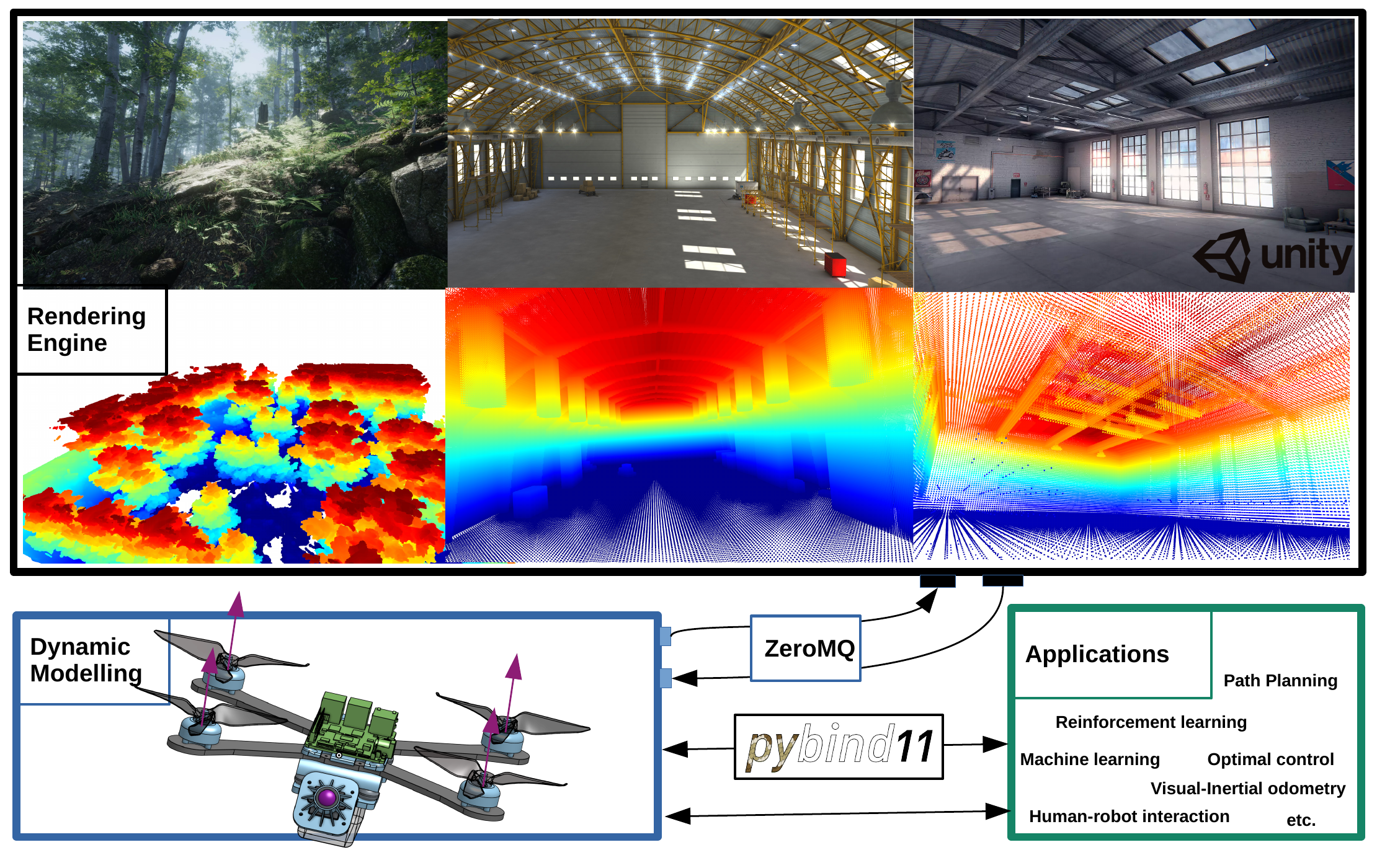}
    \caption{System overview of Flightmare.}
    \label{fig:system_diagram}
\end{figure}

Flightmare is a modular and flexible quadrotor simulator that is mainly composed of two separate components:
a photo-realistic rendering engine built with the Unity Editor and a quadrotor dynamics simulation. 
We decouple the quadrotor's dynamic modeling from the rendering engine in order to achieve fast and 
accurate dynamics simulation by making use of parallel programming~\citep{dagum1998openmp}.
Flightmare provides a flexible interface for the user to simulate different sensors in various complex
close-to-reality 3D environments.
The interface between the rendering engine and the quadrotor dynamics is implemented
using a high-performance asynchronous messaging library ZeroMQ~\footnote{https://zeromq.org}. 
In addition, we use the python wrapper~\citep{jakob2017pybind11} to implement OpenAI-Gym style interface for 
reinforcement learning tasks. 
A system overview of Flightmare is shown in Figure~\ref{fig:system_diagram}.

\subsection{Rendering Engine}
\textbf{3D Environments.}
The rendering engine of Flightmare was built with Unity, which is a popular cross-platform game engine 
and a general platform for artificial intelligence~\citep{juliani2018unity}. 
Unity enables high-fidelity graphical rendering, including realistic pre-baked or real-time lighting,
flexible combinations of different meshes, materials, shaders, and textures for 3D objects, 
skyboxes for generating realistic ambient lighting in the scene, and camera post-processing. 
Flightmare offers various high-quality 3D environments: from a simple warehouse to a complex nature forest, where
the environments are composed of high-resolution 3D models of both static and dynamic objects.
A new environment or asset can easily be created or directly purchased from the Unity Asset Store\footnote{https://assetstore.unity.com/}.
Hence, it is straightforward for the user to extend or change the environment with only very limited knowledge of the Unity Editor. 

\textbf{Sensors.}
The rendering engine provides a flexible configuration of the sensor suite. 
Currently, Flightmare can simulate RBG cameras with ground-truth depth and semantic segmentation, 
rangefinders, and collision detection between agents and their surroundings.
In particular, Flightmare allows the user to change the camera intrinsics such as field of view, focal length, and lens distortion. 
In addition, it can simulate physical effects on the camera including motion blur, lens dirt, and bloom.
An arbitrary number of sensors and their extrinsic parameters with respect to the vehicle's body frame can be defined offline or online. 
In addition, we provide a graphical user interface (GUI) as well as a C++ application programming interface (API)
for users to extract ground-truth point clouds of the environment, and then, export it as a binary polygon file format (PLY). 
The PLY file stores the ground-truth three-dimensional information about the environment by checking the occupancy of a specific point. 
The point cloud file can be read via the Open3D~\citep{open3d} library and used for path planning algorithms (see Section~\ref{sec: exp}). 

\begin{figure}[!htp]
    \centering
    \includegraphics[width=1.0\textwidth]{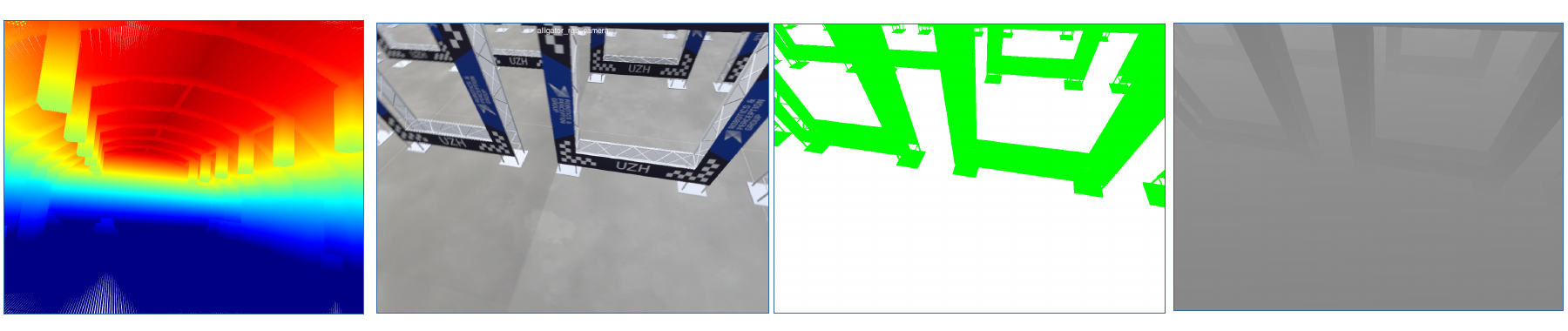}
    \caption{Screenshots of the sensor information provided by Flightmare.}
    \label{fig:sensors}
\end{figure}

\textbf{Scripts.}
Unity offers a rich and flexible scripting system via
C\#, making it possible for the user to define complex simulation tasks, such as creating graphical effects and
controlling the physical behavior of objects.
We use a list of C\# scripts for different tasks, including scene management, object control, image synthesis, sensor instantiation, and simulation. 
All scripts are developed independent of the simulated 3D environments, and hence, are easy to add to other existing Unity projects.

\subsection{Dynamic Modeling}
Flightmare provides a flexible interface to three quadrotor dynamics: a gazebo-based quadrotor dynamics~\citep{furrer2016rotors}, 
real-world dynamics, and a parallelized implementation of classical quadrotor dynamics~\citep{Faessler18ral}. 
Each dynamics serves useful purposes for different applications. 
For example, we can simulate hundreds of racing drones in parallel and collect several millions of state transitions under a minute. 
Such a parallel sampling scheme is extremely useful for large-scale reinforcement learning applications. 
The gazebo-based dynamics is slower, but more realistic thanks to high-fidelity physics engines, \emph{e.g.} Bullet. 
Finally, Flightmare offers the interface to combine real-world dynamics with photo-realistic rendering, 
similarly to what is already shown by previous work~\citep{guerra2019flightgoggles}. 

The quadrotor is modeled as a rigid body which is actuated by four motors.
We use the quadrotor dynamics that have been used for designing control algorithms in
real quadrotor experiments~\citep{Faessler17ral, Faessler18ral, kaufmann2020RSS}:
\begin{align*}
\mathbf{\dot{p}}_{WB} &= \mathbf{v}_{WB}   & 
\mathbf{\dot{v}}_{WB} &= \mathbf{q}_{WB} \odot \mathbf{c} - \mathbf{g} - \mathbf{RDR}^T\mathbf{v} \\
\mathbf{\dot{q}}_{WB} &= \frac{1}{2} \mathbf{\Lambda} ( \boldsymbol{\omega}_{B}) \cdot \mathbf{q}_{WB}   & 
\boldsymbol{\dot{\omega}}_{WB} &= \mathbf{J}^{-1} (\boldsymbol{\eta} - \boldsymbol{\omega}_{WB} \times \mathbf{J}\boldsymbol{\omega}_{WB})   
\end{align*}
where $\mathbf{p}_{WB}$ is the position, $\mathbf{v}_{WB}$ is the linear velocity of the quadrotor in the world frame~${W}$,
and $\mathbf{D} = \text{diag}(d_x, d_y, d_z)$ is a constant diagonal matrix which defines the rotor-drag coefficients, which is a linear effect in the quadrotor's velocity. 
We use a unit quaternion $\mathbf{q}_{WB}$ to represent the orientation of the quadrotor and 
use $\boldsymbol{\omega}_{WB}$ to denote the body rates in the body frame~${B}$.
Here, $\mathbf{g}=[0, 0, -g_z]^{T}$ with $g_z=\SI{9.81}{\meter\per\second\squared}$ is the gravity vector,
and $\mathbf{\Lambda} (\boldsymbol{\omega}_{B})$ is a skew-symmetric matrix.
Moreover, $\mathbf{c}=[0, 0, c]$ is the mass-normalized thrust vector. 
The conversion of single rotor thrusts $[f_1, f_2, f_3, f_4]$ to the mass-normalized thrust $c$ 
and the body torques $\boldsymbol{\eta}$ is formulated as
\begin{align}
    \boldsymbol{\eta} &= \begin{bmatrix}
    \frac{l}{\sqrt{2}}(f_1 - f_2 - f_3 + f_4)  \\ 
    \frac{l}{\sqrt{2}}(-f_1 - f_2 + f_3 + f_4) \\
    \kappa f_1 - \kappa f_2 + \kappa f_3 - \kappa f_4 \end{bmatrix} \\
   c &= (f_1 + f_2 + f_3 + f_4) / m
\end{align}
where $m$ is the quadrotor's mass and $l$ is the arm length.
We model the dynamics of a single rotor thrusts as first-order systems
$\dot{f} =(f_\text{des} - f)/ \alpha$ where $\alpha$ is the time-delay constant.
We implemented both Euler and 4th-order Runge-Kutta methods for integrating the dynamic equations.
In addition, we simulate the inertial measurement unit (IMU) which provides acceleration and angular rate
measurements of the vehicle.
The quadrotor can be controlled in two different modes: the body-rate mode and the rotor-thrusts mode. 
In the body-rate control, we implemented a low-level controller~\citep{faessler2015automatic} for tracking 
the desired body rates, in which the low-level controller generates desired rotor thrusts for each motor.

\section{Experiments}
\label{sec: exp}
We design our experiments to answer the following questions. 
Why do we need to decouple the dynamics simulation from the rendering engine? 
What are the simulation speeds of the quadrotor dynamics and the rendering engine?
How can we use the simulator for robot learning?
What other applications can we use the simulator for?

\subsection{Simulation Speed}
\begin{figure}[!htp]
\centering
\begin{subfigure}{.5\textwidth}
    \centering
    \includegraphics[width=\textwidth]{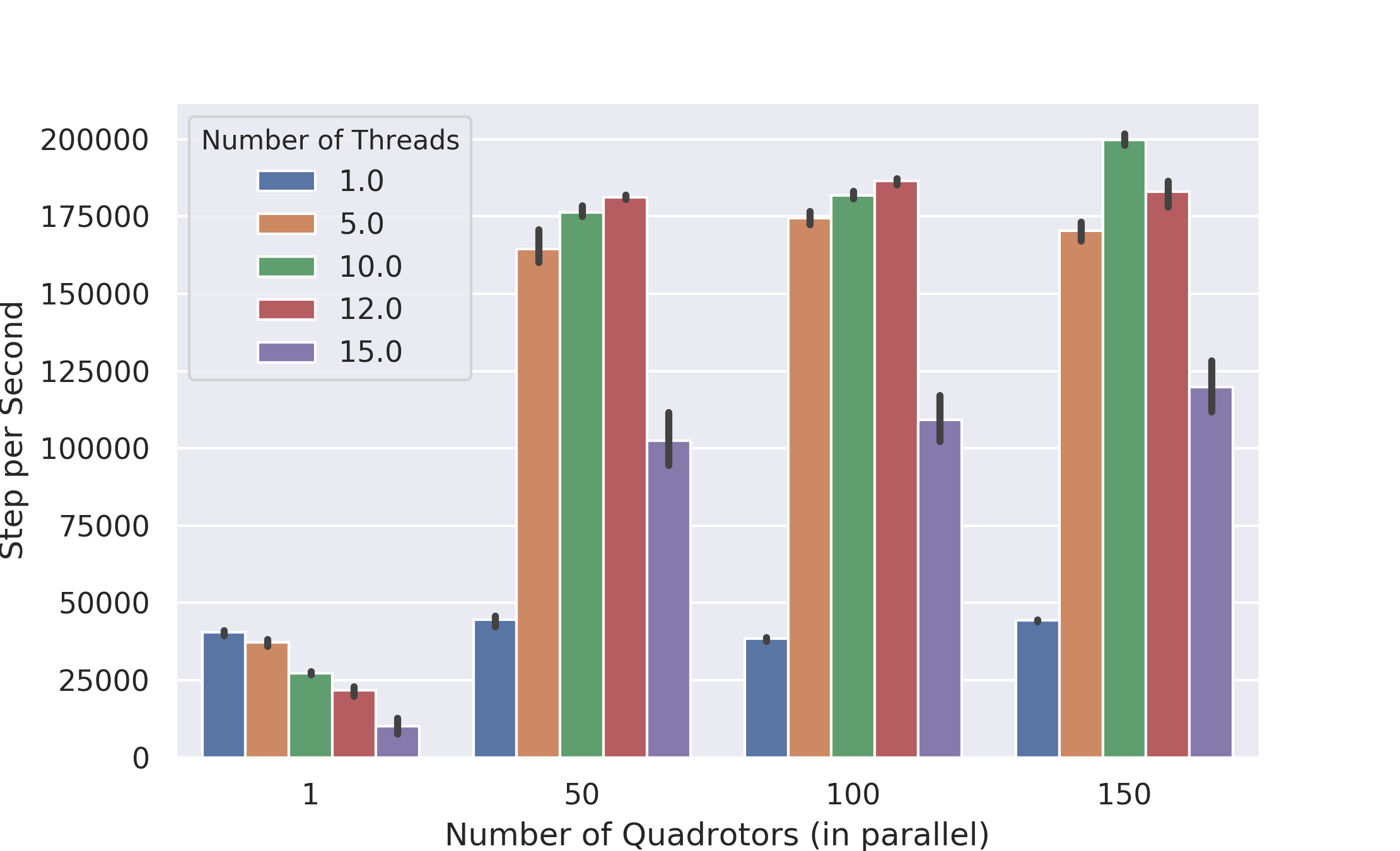}
    \label{fig:sim_dynamics_performance}
\end{subfigure}%
\begin{subfigure}{.5\textwidth}
    \centering
    \includegraphics[width=\textwidth]{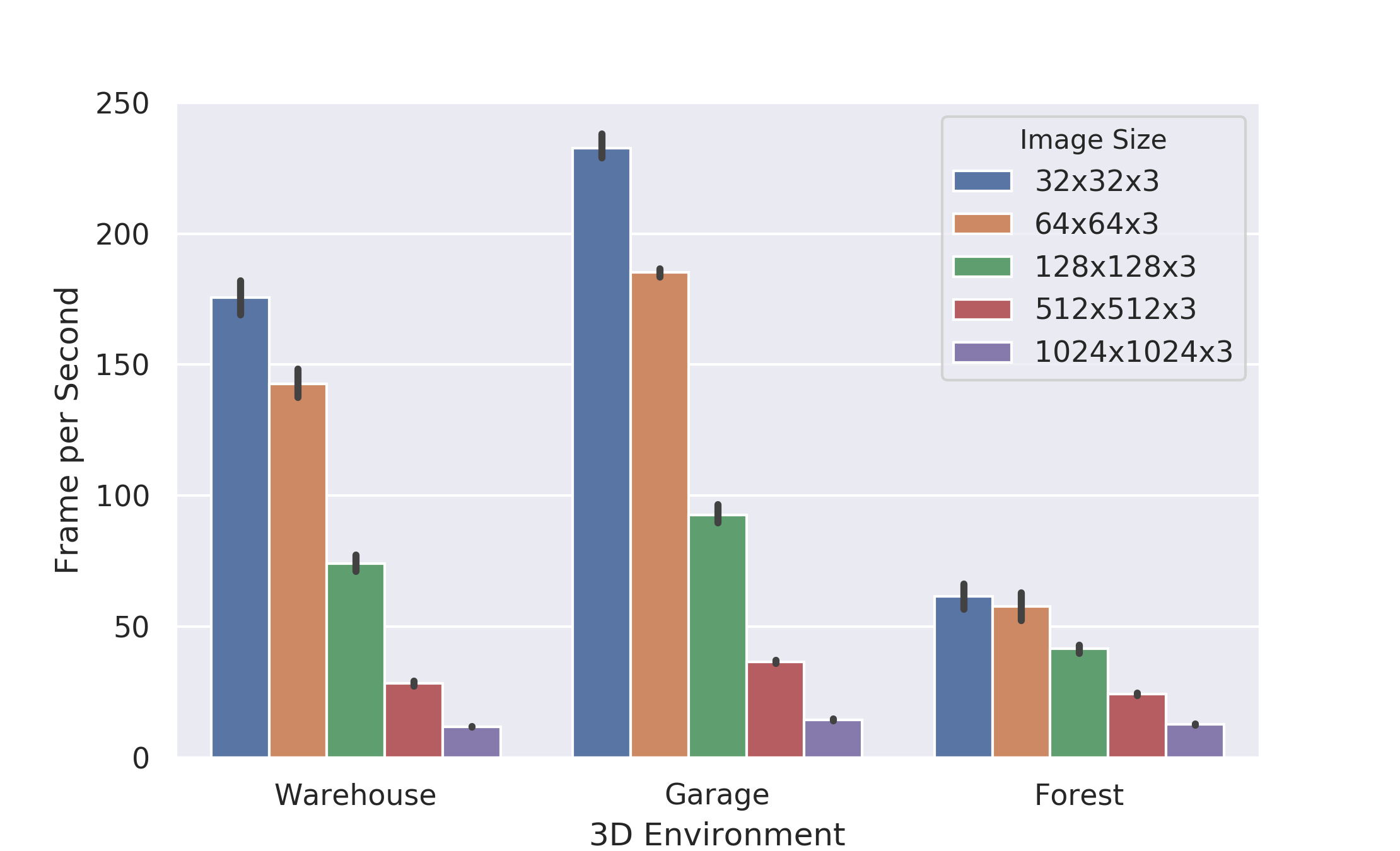}
    \label{fig:sim_rendering_performance}
\end{subfigure}
\caption{Simulation performance when using Python API to interact with the simulation. 
    \textbf{Left:} Quadrotor dynamics simulation speed with different number of simulated quadrotor in parallel and different number of threads.
    \textbf{Right:} A comparison of RGB image rendering speed in different environments with different image sizes.}
\label{fig:sim_performance}
\end{figure}
We evaluate the simulation speed of  Flightmare using a laptop with a 12-core Intel(R) Core(TM) i7-8850H CPU at up to 2.60GHz. 
The evaluation result is shown in Figure~\ref{fig:sim_performance}.
We first evaluate the speed of quadrotor dynamics by simulating multiple quadrotors in parallel using multiple CPU threads.
Flightmare is extremely fast---when simulating 150 quadrotors in parallel with randomly sampled actions, 
it achieves over 200,000 steps per second on the CPU. 
It allows users to collect several millions of samples under 
a minute.
For example, we achieved an average sampling rate of around 2 million samples per minute when 
using a fully-connected multilayer perceptron~(MLP) with two hidden layers of 128 units in the sampling loop.
Since we use a laptop that has a maximum of 24 threads in total, increasing the number of threads above a certain threshold, e.g., 15 threads, 
for parallelization results in a performance drop. 
Besides, using multiple threads for a single quadrotor simulation will add additional computational and memory costs, and hence, can result in lower sampling rates. 
We test the RGB image rendering speed of simulated cameras using the Unity application,
in which the simulated 3D scenes range from a lightweight \textit{Garage} environment to a complex
\textit{Nature Forest} environment. 
Flightmare can offer greater than real-time rendering speed by taking advantages of the parallelization scheme and
a flexible API for configuring an arbitrary number of simulated cameras. 
For example, we test the rendering speed (frames per second) with 5 simulated quadrotors in parallel, each quadrotor is attached with an RGB camera (see Figure~\ref{fig:sim_performance}). 
We achieve frame rates of up to 230~Hz for the smaller image size in the \textit{Garage} environment.

\subsection{Learning a Sensorimotor Policy for Quadrotor Control}
\begin{figure}[!htp]
\centering
\includegraphics[width=1.0\textwidth]{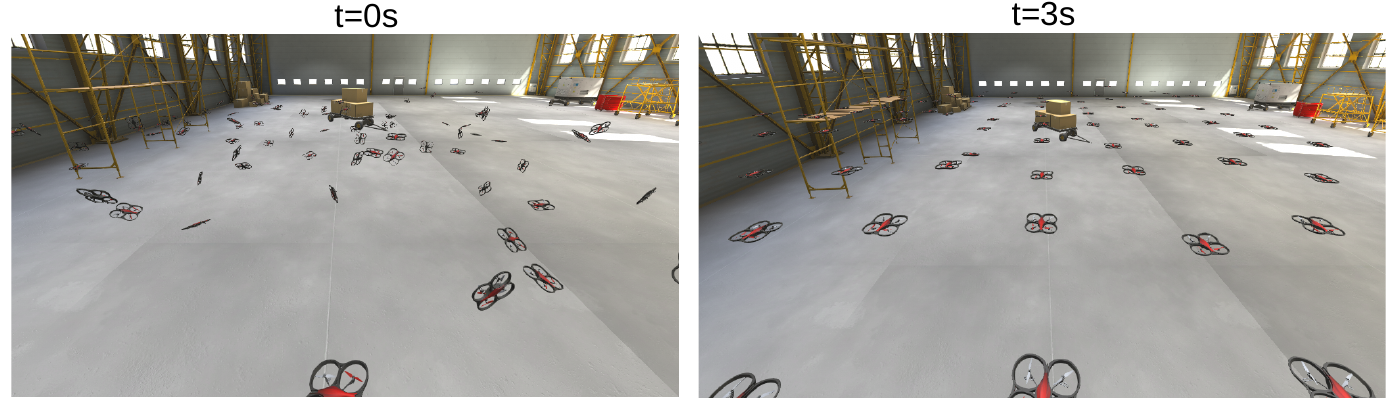}
\caption{Control of 100 quadrotors in parallel using the learned sensorimotor policy.}
\label{fig:rl_drone_control_demo}
\end{figure}

\begin{figure}[!htp]
\centering
\begin{subfigure}{.333\textwidth}
    \centering
\includegraphics[width=1.0\textwidth]{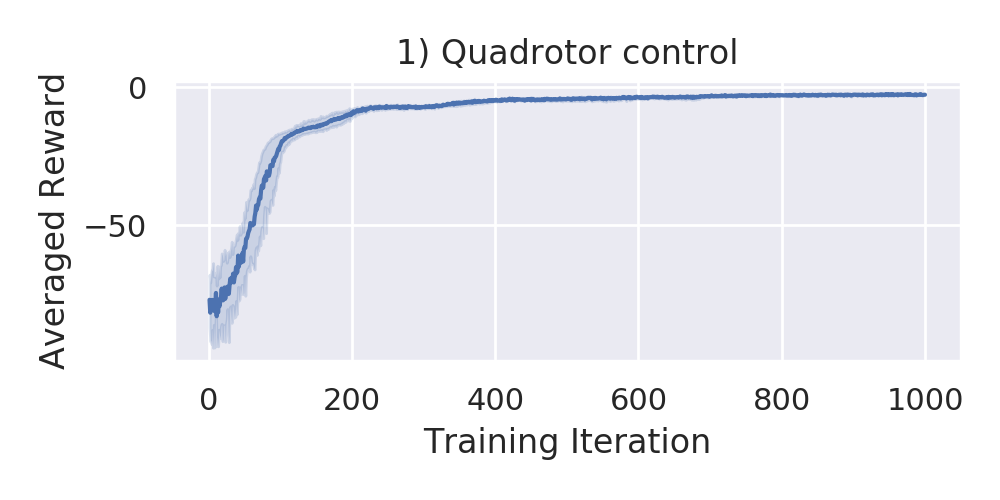}
    \label{fig:rl_quadrotor_control}
\end{subfigure}%
\begin{subfigure}{.333\textwidth}
    \centering
\includegraphics[width=1.0\textwidth]{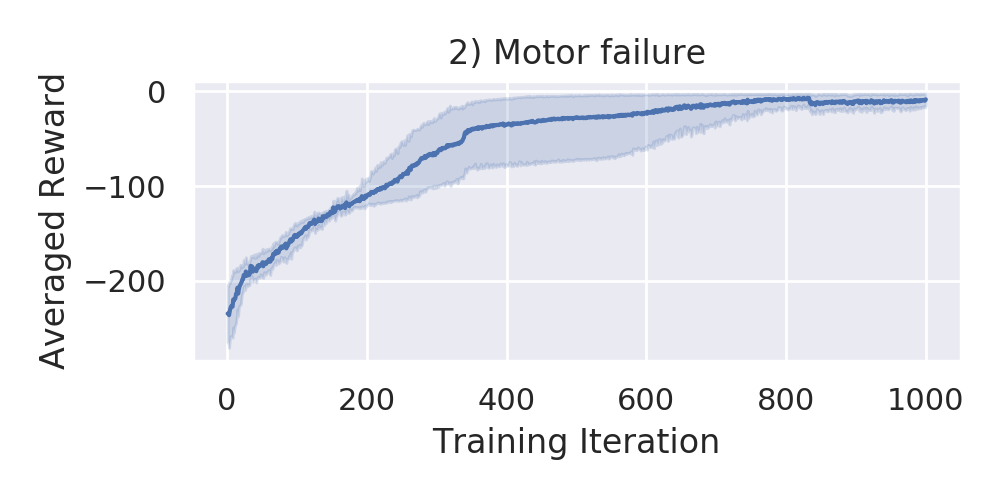}
    \label{fig:rl_quadrotor_control_motor_failure}
\end{subfigure}%
\begin{subfigure}{.333\textwidth}
    \centering
    \includegraphics[width=1.0\textwidth]{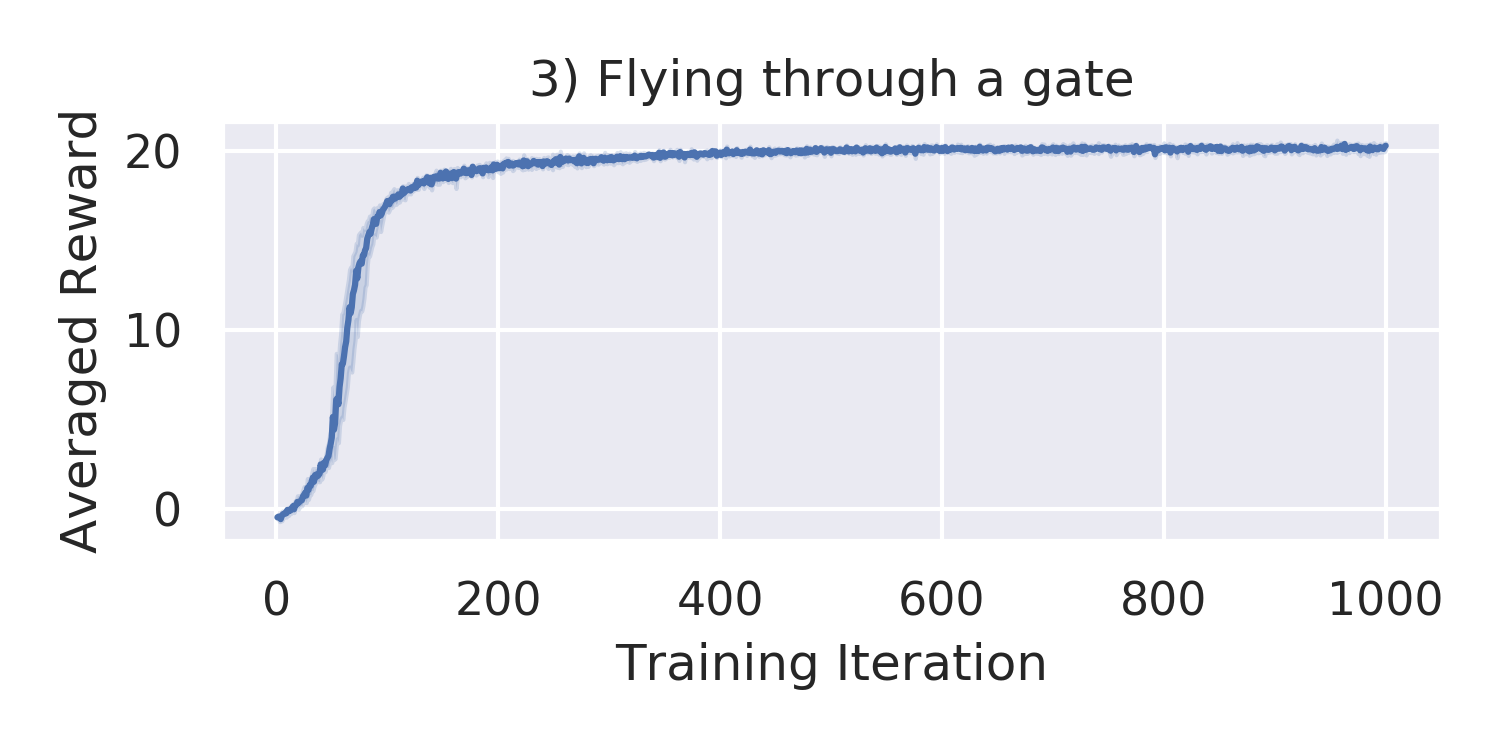}
    \label{fig:rl_quadrotor_control_racing}
\end{subfigure}
\caption{Learning curve of 3 different tasks using PPO. The shaded region indicates the standard deviation over 5 different random seeds. }
\label{fig:rl_rewards}
\end{figure}

Flightmare provides several example tasks as well as OpenAI gym-style~\citep{openai_gym} wrappers for reinforcement learning.
Those gym wrappers give researchers a user-friendly interface for the interaction between Flightmare
and existing RL baselines designed around the gym interface. 
These tasks are designed to be both useful for benchmarking RL algorithms as well as templates for solving more complex problems.
We list the RL tasks as well as their input states and output control actions in Table~\ref{tab:rl_tasks}.
It includes the following tasks: 
1) stabilizing a quadrotor from randomly initialized poses (similar to~\citep{rl_quadrotor});
2) stabilizing a quadrotor from randomly initialized poses under a single motor failure;
3) controlling a quadrotor to fly through static gates as fast as possible.
These tasks feature interesting research problems in both robotics and reinforcement learning, such as 
quadrotor control using neural networks and learning a time-optimal controller in drone racing. 
We train neural network controllers for each task using the Proximal Policy Optimization~(PPO) algorithm~\citep{schulman2017proximal}
and the OpenAI stable-baselines implementation~\citep{stable-baselines}. 
During training, we simulate 100 quadrotors in parallel for trajectory sampling and collect in total 25 million time steps for each task.
The learning curves for these tasks are reported in Figure~\ref{fig:rl_rewards}.
Figure~\ref{fig:rl_drone_control_demo} shows
screenshots of controlling 100 quadrotors in parallel using a single pre-trained neural network controller.

\begin{table}[!htp]
    \centering
    \begin{tabular}{l|c|c}
      \hline
      \textbf{Task} & \textbf{Input Observations} & \textbf{Output Actions}\\
      \hline
      1) Quadrotor control & $[\mathbf{p}, \boldsymbol{\theta}, \mathbf{v}]$, dim=10 & $[c, \omega_x, \omega_y, \omega_z]$, dim=4 \\
      2) Quadrotor control under motor failure & $[\mathbf{p}, \boldsymbol{\theta}, \mathbf{v}, \boldsymbol{\omega}]$, dim=12 &  $[f_1, f_2, f_3]$, dim=3 \\
      3) Flying through a gate & $[\mathbf{p}, \boldsymbol{\theta}, \mathbf{v}, \boldsymbol{\omega}, \mathbf{p}_\text{gate}, \boldsymbol{\theta}_\text{gate}]$, dim=18 &  $[f_1, f_2, f_3, f_4]$, dim=4 \\
    \end{tabular}
    \caption{Example tasks for reinforcement learning.}
    \label{tab:rl_tasks}
\end{table}

\subsection{Point Cloud and Path Planning}
\begin{figure}[!htp]
    \centering
    \includegraphics[width=1.0\textwidth]{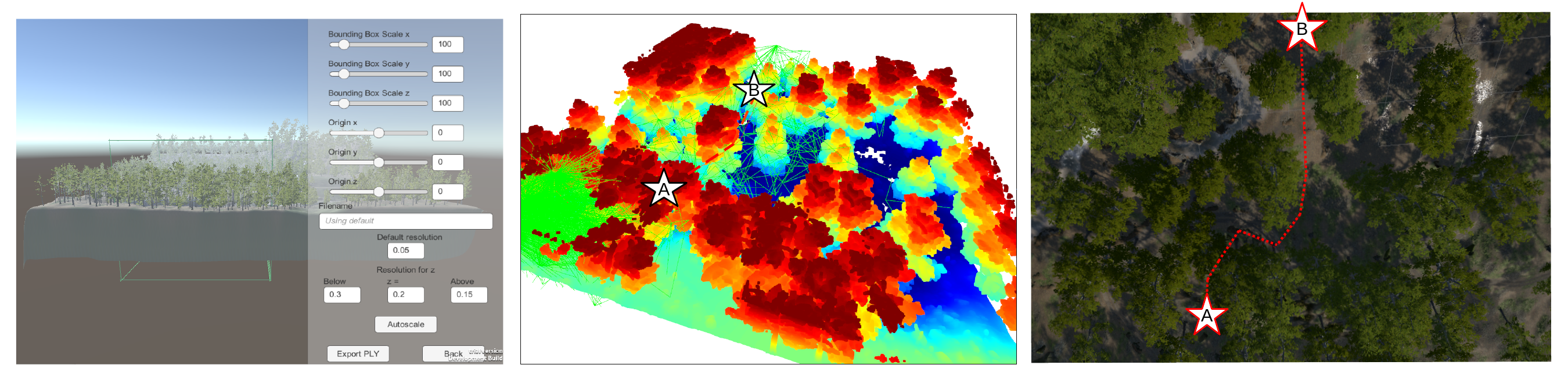}
    \caption{ \textbf{Left:} User interface for point cloud extraction. 
    \textbf{Middle:} A visualization of the extracted point cloud in the \textit{Forest} environment, all collision-free path (green line) found by OMPL,
    and solved shortest path (red line) between A and B. 
    \textbf{Right:} A bird view of the shortest path in the forest environment.}
    \label{fig:path_planning}
\end{figure}

Existing simulators don't provide an efficient API to access 3D information of the environment. 
However, such information is needed by a large class of algorithms, e.g. path-planning.
To foster research in this direction, Flightmare provides an interface to export the 3D information of the full environment (or a region of it) as point cloud with any desired resolution (configurable via the UI, Figure~\ref{fig:path_planning}).
The point-cloud is saved in the binary polygon file format (PLY), given its compatibility with the open-source library Open3D~\citep{open3d}.
We show an example of a generated point-cloud in Figure~\ref{fig:path_planning}, where we illustrate a section of the complex \textit{Nature Forest} environment.
This point-cloud is $\SI{100}{\meter}\times\SI{100}{\meter}\times\SI{30}{\meter}$ with a resolution of $\SI{0.1}{\meter}$ and contains detailed 3D structure information
of the forest, such as clusters of small tree branches and leaves. 
As an example application, we intend to compute the shortest collision-free path between two points: from point A to point B.
We run the Open Motion Planning Library~(OMPL)~\citep{sucan2012the-open-motion-planning-library} on the point-cloud extracted from the forest with a 
default solver for path-planning.
In spite of the complexity of the environment, the solver finds a solution within 1.0 second.


\subsection{Virtual Reality and Safe Human-Robot Interaction}
\begin{figure}[!htp]
\centering
\includegraphics[width=1.0\textwidth]{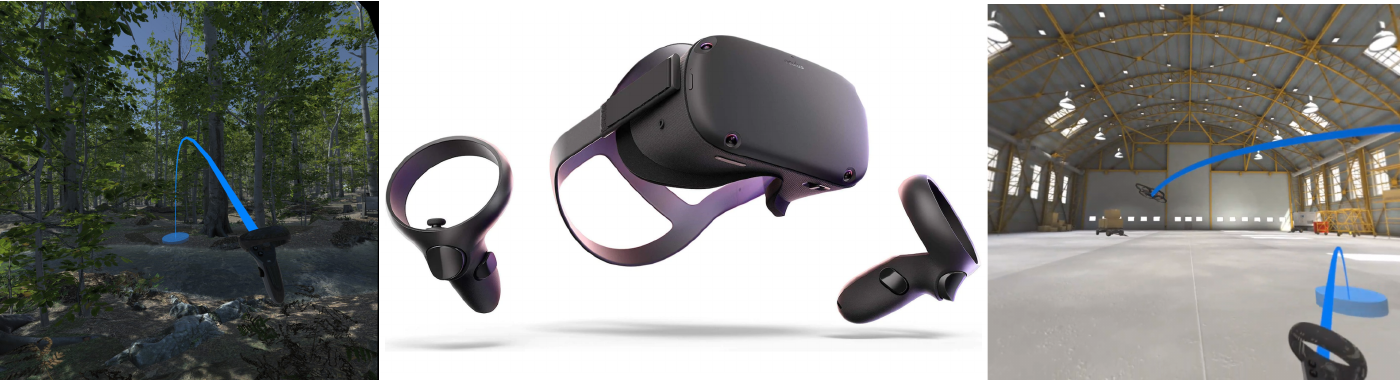}
\caption{The user can directly interact with the simulator via a virtual-reality headset.}
\label{fig:VR}
\end{figure}

As autonomous flying robots and the consumer drone market flourish, safe collocated human-drone interactions are 
becoming increasingly important.
Compared to ground robotics, the high-speed motion of drones poses different challenges 
in the recent research area of human-robot interaction (HRI)~\citep{human_drone}.
To open Flightmare for this new research community, we integrate our simulator with the popular Oculus virtual reality offset (Figure~\ref{fig:VR}).
This opportunity offers a favorable alternative to real-world experiments for HRI. Indeed, it allows interaction with the drone under a large set of configurations, included some extreme, potentially dangerous, cases.
%
Other applications that could potentially benefit from the virtual reality feature of Flightmare are human-aware robot navigation, safe drone landing in cities, and safe pilot training.

\subsection{Other Applications}
Flightmare is not limited to the aforementioned applications, and it can be used to several other applications.
Given its ability to simulate hundreds of quadrotors in parallel, our simulator can be used to study the implications of large-scale multi-robot systems.
In addition, given its flexible structure and the availability of hundreds of simulated environments through the Unity Store, we believe that Flightmare can be extremely useful for testing odometry and simultaneous localization and mapping~(SLAM) systems. 
In the context of robot learning, Flightmare can also be used to learn deep sensorimotor policies via imitation learning~\citep{kaufmann2020RSS}.

\section{Conclusion and Discussion}
\label{sec: conclusion}
Being optimized for specific tasks or features, \emph{e.g.} photo-realistic rendering, currently available quadrotor simulators have been developed with a rigid structure.
However, this design choice constrains the set of applications these simulators can be used for.
Specifically, it cuts out the possibility of letting the user, or an automated algorithm, fine-tune the simulator to the task at hand.
Our work moves away from this rigid paradigm and proposes a flexible modular structure that empowers the users with full control of the simulation characteristics.

According to this idea, we design a novel simulator for quadrotors: Flightmare.
In Flightmare, physics modeling and visual sensors rendering are managed by two independent blocks.
Physics simulation can be adapted to follow the robot dynamics with any degree of accuracy, from the simplest point-mass to real-world quadrotor dynamics.
Similarly, rendering can be configured to accommodate the different needs of the users and ranges from a very fast, but simplistic, version to a more photo-realistic, but slower, configuration.
In addition, Flightmare possesses several favorable features with respect to previous work: a large sensor suite fitting to the majority of robotics and machine learning applications, an API to efficiently simulate multiple quadrotors in parallel and train controllers with reinforcement learning, and the possibility to interact with the simulator via a virtual reality headset.

Our work opens up several opportunities for future work.
On the systemic side, it would be interesting to apply the proposed flexible design to other robots, \emph{e.g.} manipulators or ground vehicles.
Our experimental finding, currently limited to the context of quadrotors, will most likely generalize to these other platforms.
On the algorithmic side, it would be interesting to develop new methods to optimize the simulation to the task in an end-to-end fashion.
Eventually, not only does Flightmare foster new research opportunities for autonomous navigation, but also for the human-robot interaction community.




\clearpage
\acknowledgments{This work was supported by the National Centre of Competence in Research (NCCR) Robotics through the Swiss National Science Foundation, the SNSF-ERC Starting Grant, and the European Union’s Horizon2020 research
and innovation program through the AERIAL-CORE project (H2020-2019-871479). We would like to thank Philipp Foehn for structuring the C++ project and helping with the open source implementation of Flightmare.}


\bibliography{ref} 

\newpage
\section*{Supplementary Material}
In this section, we describe the experiment details that are used in the reinforcement learning tasks of learning sensorimotor policies for quadrotor control.

\subsection{Quadrotor Control}
In this task, we aim to learn a neural network policy for quadrotor stabilization under random initialization, similar to~\citep{rl_quadrotor}. At the beginning of each episode, the quadrotor's state $\mathbf{s} = [\mathbf{p}, \boldsymbol{\theta}, \mathbf{v}]$ is randomly initialized, where $\mathbf{p}$ is the position vector, $\boldsymbol{\theta}$ represents the orientation using Euler angles, and $\mathbf{v}$ is the linear velocity vector. The goal is to stabilize the quadrotor in the goal state $\mathbf{s}_\text{target} = [\mathbf{p}_\text{target}, \boldsymbol{\theta}_\text{target}, \mathbf{v}_\text{target}]$, where $\boldsymbol{\theta}_\text{target}=[0, 0, 0]$ are Euler angles in the hovering mode and $\mathbf{v}_\text{target}=[0, 0, 0]$ are the target linear velocities.
We train a multilayer perceptron (MLP) using the Proximal Policy Optimization~(PPO)~\citep{schulman2017proximal}, where the output of the MLP is the body-rate control command $[c, \omega_x, \omega_y, \omega_z]$. The mass normalized thrust $c$ and desired body rates $\boldsymbol{\omega} = [\omega_x, \omega_y, \omega_z]$ are outputted to a low-level controller. We define a reward function for the reinforcement learning 

\begin{equation}
r_t = - (c_1\| \mathbf{p} - \mathbf{p}_\text{target}\| + c_2\| \boldsymbol{\theta} - \boldsymbol{\theta}_\text{target}\| + c_3\| \mathbf{v} - \mathbf{v}_\text{target}\|) 
\label{eq:rew_ctl}
\end{equation}
where $c_1 = 2*10^{-3}, c_2 = 2*10^{-3}, c_3 = 2*10^{-4}$ are the weights for each reward component. 
We use a fixed episode length of $\SI{5}{\second}$ and a discretization of $d_t = \SI{0.02}{\second}$ for the dynamics simulation. 

\subsection{Quadrotor Control Under Motor Failure}
Controlling a quadrotor under motor failure is a more challenging problem. This example describes a new application of using reinforcement learning for quadrotor control despite the loss of one propeller. Similar to the previous example, we initialize the quadrotor with random poses in the air and intend to stabilize it with an MLP. The input state to the MLP is denoted as $\mathbf{s} = [\mathbf{p}, \boldsymbol{\theta}, \mathbf{v}, \boldsymbol{\omega}]$ of dimension $\text{dim}=12$, where $\boldsymbol{\omega}$ denotes the body rates in the vehicle's body frame. 
The neural network outputs the control commands of 3 motor thrusts acting directly on the vehicle. 
The goal is to stabilize the quadrotor in the goal state $\mathbf{s}_\text{goal} = [\mathbf{p}_\text{target}, \boldsymbol{\theta}_\text{target}, \mathbf{v}_\text{target}, \boldsymbol{\omega}_\text{target}]$, in which $\boldsymbol{\theta}_\text{target}=\mathbf{0}$, $\mathbf{v}_\text{target}=\mathbf{0}$, and $\boldsymbol{\omega}_\text{target} = \mathbf{0}$. 
We design the reward function as 
\begin{equation*}
r_t^{1} = - (c_1\| \mathbf{p} - \mathbf{p}_\text{target}\| + c_2\| \boldsymbol{\theta}_{x,y}\| + c_3\| \mathbf{v}\|+ c_4\| \boldsymbol{\omega}_{x,y}\|) 
\end{equation*}
where $c_1 = 2*10^{-3}, c_2 = 2*10^{-3}, c_3 = 2*10^{-4}, c_4 = 2*10^{-4}$ are the weights for each reward component. 
Here, we use zero weights for both the yaw angle $\theta_z$ and the body rate in yaw $\omega_z$, due to the loss of a propeller. 

\subsection{Flying Through a Gate}
Flying a quadrotor through a static gate generally requires a high-level trajectory generator 
as well as a low-level controller for tracking the trajectory~\citep{foehn2020alphapilot}. 
In this example, We show it is possible to directly learn an end-to-end controller for this 
task while forgoing the need for a high-level trajectory generator. 
During training, we randomly sample the vehicle's position in front of a gate and keep the orientation to the world's origin. 
The goal is to fly through the gate in the  $x$-axis while reaching a target hovering state $\mathbf{s}_\text{target}$ behind the gate. 
The gate is defined as a circle of radius $r=\SI{1.0}{\meter}$. 
We define a boolean expression to decide whether or not an episode is terminated.
For example, an episode is terminated when the quadrotor is hitting the gate or passing through an external region outside of the gate. 
The observation vector is a concatenation of the quadrotor's full state $\mathbf{s} = [\mathbf{p}, \boldsymbol{\theta}, \mathbf{v}, \boldsymbol{\omega}]$ and the gate's pose  $\mathbf{s}_\text{gate} =  [\mathbf{p}_\text{gate}, \boldsymbol{\theta}_\text{gate} ]$, where $\mathbf{p}_\text{gate}$ is the position vector and $\boldsymbol{\theta}_\text{gate}$ is the orientation vector of the gate in the world frame. The output control command are 4 motor thrusts $[f_1, f_2, f_3, f_4]$ acting directly on the quadrotor. 
We design the reward function as
\begin{equation}
r_t = \begin{cases}
   r_{\text{goal}} + 0.1, & \text{if not hitting the gate or the ground} \\
    -0.1,              & \text{otherwise}
\end{cases}
\end{equation}
where $r_{\text{goal}}$ has the same definition as~Equation~(\ref{eq:rew_ctl}). 
We use a positive reward $0.1$ to encourage the policy not to hit the gate and the ground. 

    

\end{document}